%% file: latex/acl_latex.tex
\pdfoutput=1

\documentclass[11pt]{article}

\usepackage[final]{acl}

\usepackage{times}
\usepackage{latexsym}

\usepackage[T1]{fontenc}

\usepackage{amsmath}

\usepackage{caption}
\usepackage{cleveref}

\crefname{figure}{Fig.}{Figs.}

\crefname{appendix}{App.}{App.}
\usepackage{subcaption}
\usepackage{tabularx}
\usepackage{booktabs}
\usepackage{multirow}
\usepackage[table]{colortbl}
\usepackage{array}
\usepackage{adjustbox}
\usepackage{hyperref}
\usepackage{makecell}

\usepackage{afterpage}
\usepackage{pgfplots}
\pgfplotsset{compat=1.18} 
\usepackage{pgfplotstable}
\usepgfplotslibrary{groupplots}
\usepackage{xcolor}

\usepackage{listings}

\lstdefinelanguage{json}{
  basicstyle=\ttfamily\small\color{black},
  stringstyle=\ttfamily\small\color{black},
  commentstyle=\ttfamily\small\color{black},
  numberstyle=\ttfamily\small\color{black},
  keywordstyle=\ttfamily\small\color{black},
  morestring=[b]",
  breaklines=true,
}

\usepackage{soul}
\usepackage{enumitem}  
\usepackage{float}
\usepackage{placeins} 

\definecolor{cb-burgundy}   {RGB}{146,   0,   0}
\definecolor{purple}{HTML}{332288}
\definecolor{green}{HTML}{117733}
\definecolor{teal}{HTML}{44AA99}
\definecolor{lightblue}{HTML}{88CCEE}
\definecolor{beige}{HTML}{DDCC77}
\definecolor{salmon}{HTML}{CC6677}
\definecolor{pink}{HTML}{AA4499}
\definecolor{maroon}{HTML}{882255}
\definecolor{cb-lilac}{RGB}{182, 109, 255}
\definecolor{softeryellow}{HTML}{FFFFCC}

\usepackage{soul}

\newcommand{\markerrname}[1]{{\color{green} \ul{\textbf{#1\textsuperscript{N}}}}}
\newcommand{\markerrnumber}[1]{{\color{purple} \ul{\textbf{#1\textsuperscript{U}}}}}
\newcommand{\markerrwordobj}[1]{{\color{salmon} \ul{\textbf{#1\textsuperscript{WO}}}}}
\newcommand{\markerrwordsub}[1]{{\color{cb-lilac} \ul{\textbf{#1\textsuperscript{WS}}}}}
\newcommand{\markerrcontext}[1]{{\color{cb-burgundy} \ul{\textbf{#1\textsuperscript{C}}}}}
\newcommand{\markerrother}[1]{{\color{lightblue} \ul{\textbf{#1\textsuperscript{O}}}}}
\newcommand{\markerrnotcheckable}[1]{{\color{maroon} \ul{\textbf{#1\textsuperscript{X}}}}}

\definecolor{red}{HTML}{DD0000}

\usepackage{comment}
\usepackage{booktabs}
\usepackage{todonotes}
\usepackage{algorithm}
\usepackage{listings}
\definecolor{red}{HTML}{DD0000}

\newcommand{\lineacross}{\rule{\linewidth}{1pt}}

\newcommand{\figline}[1][1.0pt]{%
  \par\noindent{\color{black}\rule{\linewidth}{#1}}\par
}

\usepackage{array}
\newcolumntype{P}[1]{>{\centering\arraybackslash}p{#1}}

\usepackage[page]{appendix}

\usepackage{subcaption}

\usepackage[utf8]{inputenc}

\usepackage{microtype}

\usepackage{inconsolata}

\usepackage{graphicx}

\newcommand{\llamafam}{Llama}
\newcommand{\qwenfam}{Qwen}

\newcommand{\LlamaVer}[2]{\llamafam-#1-#2}  
\newcommand{\QwenVer}[1]{\qwenfam#1}        

\newcommand{\na}{\texttt{N/A}}

\title{Input Matters: Evaluating Input Structure's Impact on LLM Summaries of Sports Play-by-Play}

\author{Barkavi Sundararajan \and Somayajulu Sripada \and Ehud Reiter\\
  Department of Computing Science, University of Aberdeen \\
  \texttt{\{b.sundararajan.21, yaji.sripada, e.reiter\}}{@abdn.ac.uk} \\}

\begin{document}
\maketitle
\begin{abstract}
A major concern when deploying LLMs in accuracy-critical domains such as sports reporting is that the generated text may not faithfully reflect the input data. We quantify how input structure affects hallucinations and other factual errors in LLM-generated summaries of NBA play-by-play data, across three formats: row-structured, JSON and unstructured. We manually annotated 3,312 factual errors across 180 game summaries produced by two models, \LlamaVer{3.1}{70B} and \QwenVer{2.5-72B}. Input structure has a strong effect: JSON input reduces error rates by 69\% for \llamafam{} and 65\% for \qwenfam{} compared to unstructured input, while row-structured input reduces errors by 54\% for \llamafam{} and 51\% for \qwenfam{}. A two-way repeated-measures ANOVA shows that input structure accounts for over 80\% of the variance in error rates, with Tukey HSD post hoc tests confirming statistically significant differences between all input formats.




\end{abstract}


\input{latex/tables/json_structure}
\section{Introduction}
\input{latex/tables/row_structure}
Despite recent advancements, large language models (LLMs) are known to produce factually incorrect output \citep{jacovi2025factsgroundingleaderboardbenchmarking}. These errors occur frequently enough to make LLM-based NLG applications unacceptable in accuracy-critical domains, unless deployed in a human-in-the-loop setting where mistakes can be corrected. Sports reporting is one such domain, and in this paper, we investigate the factual accuracy of LLMs for generating NBA game summaries. This is a challenging text generation task because play-by-play data is highly detailed, encoding \emph{entities} (teams, players), \emph{events/actions} (e.g., jump shot, layup, free throw, offensive rebound), \emph{relationships} (e.g., \emph{player}~--~\emph{attempts}~--~\emph{shot}; \emph{player}~--~\emph{commits}~--~\emph{foul}), and \emph{attributes} (e.g., scores, time, period, distance). 


We hypothesise that structuring the complex play-by-play data is important for producing factually accurate summaries, whereas unstructured input (\emph{garbage in}) yields more factual errors (\emph{garbage out}). To empirically study the impact of input structuring on model output, we evaluate three input formats: unstructured (natural play-by-play descriptions; see \Cref{tab:unstructured_ot1}), row-structured tabular format (see \Cref{tab:row_example}), and hierarchical JSON. \Cref{fig:play_json} shows a play-by-play event represented as a JSON object. 

Recent work by \citet{10.1145/3616855.3635752} shows that different tables define structure and formatting in distinct ways, which can influence how LLMs interpret and process tabular inputs. While their focus was on QA and relatively simple generation tasks \citep{parikh-etal-2020-totto}, we study a more complex scenario: summarising NBA play-by-play data, which is event-driven, temporally ordered, and densely factual with over 450 rows and 13 columns. This makes it a more challenging and underexplored setting for investigating the impact of input structure on LLM-based summarisation.

To investigate this, we evaluate two open-source LLMs, \LlamaVer{3.1}{70B} \citep{grattafiori2024llama3herdmodels} and \QwenVer{2.5-72B} \citep{Yang2024Qwen25TR}. We use a manual annotation protocol adapted from \citet{10.1016/j.csl.2023.101482} and \citet{sundararajan-etal-2024-improving} to assess how input structure affects factual errors in generated summaries.

The key findings from our study are as follows:

\begin{itemize}[leftmargin=1.2em, itemsep=0.2ex, topsep=0.2ex, parsep=0pt, partopsep=0pt]
  \item A two-way repeated-measures ANOVA shows that structured inputs significantly reduce error rates compared with unstructured text, with JSON providing the strongest improvements (p~$<\!0.05$).
  \item Error-type analysis shows that \LlamaVer{3.1}{70B} produces more \emph{Number} errors, whereas \QwenVer{2.5-72B} produces more \emph{Name} and \emph{Word-subjective} errors. Across both models, \emph{Word-objective} errors dominate; common patterns are
event/action substitutions (e.g., mistakenly produces \emph{a pair of free throws} instead of \emph{a layup}).

\end{itemize}

\section{Related Work}
\label{sec: related_work} 

Prior work on data-to-text generation for basketball \citep{wiseman-etal-2017-challenges, puduppully-et-al-2019-planning, thomson-etal-2020-sportsett} has used hierarchical encoder models and related architectures to generate summaries from box scores (aggregated player and team statistics), yet consistent factual errors have been reported \citep{10.1016/j.csl.2023.101482}. The advent of long-context LLMs (up to 128k tokens) \citep{wu2024extending} has enabled studies on richer inputs; for example, \citet{hu-etal-2024-reasoning} use NBA play-by-play data to compute player and team points. Their work focuses on reasoning and aggregation over play-by-play data, supporting the value of using a single rich dataset for controlled analysis. In contrast, we generate paragraph-style summaries from complete play-by-play data and compare three input structures: unstructured input with 13,000 tokens, row-structured input with 28,000 tokens, and JSON input with 70,000 to 80,000 tokens, producing summaries of 450 to 500 words.

\Citet{maynez-etal-2020-faithfulness} demonstrated that automatic metrics are insufficient for measuring hallucinations. Building on \citet{10.1016/j.csl.2023.101482} and \citet{sundararajan-etal-2024-improving}, we adapted their manual error annotation protocols to study factual errors at a granular level, making minor task-specific adjustments for the NBA play-by-play data to evaluate the errors in our generated summaries.

\section{Experimental Setup}
\label{sec: experiment}

\subsection{Game Selection Criteria}
\label{subsec: selection} 
\Cref{fig:strata_doublebar} shows our stratified random sampling based on the final point difference (\emph{absolute score difference between two teams}). We collected all NBA games from Dec.~2024--Jan.~2025 to limit potential data contamination \citep{balloccu-etal-2024-leak}, partitioned the games into four strata: ($\leq 3$, 4--9, 10--19, $\geq 20$ point differences), and randomly sampled within each stratum to obtain a balanced representation of games, yielding 30 games in total.

\subsection{Input Representation and Preprocessing}
\label{subsec: preprocess} 
We processed raw NBA play-by-play logs from Basketball Reference\footnote{\url{https://www.basketball-reference.com/boxscores/pbp/202501010DET.html}} to generate three types of input structures: row-structured, JSON, and unstructured. We performed light cleaning and assigned clear column names, then decomposed each play description into atomic fields (e.g., primary/secondary player, team, action and event type, outcome, distance, time remaining, period, score), applying regular expressions and task-specific rules to normalise values. From these atomic fields, we produced a row-structured table with one row per play (\cref{tab:row_example}). In parallel, we encoded a hierarchical JSON that nests entities (player, action, team) and their attributes under each play \cref{fig:play_json}.

For the unstructured variant, we preserved the original natural-language descriptions for both teams and applied only lightweight regex-based cleanup (removing extra spaces around player names and actions) without altering the underlying structure. We release the structured play-by-play datasets (row-structured and JSON)\footnote{\url{https://github.com/BarkaviSJ/nba-input-matters}}.
    
    

\input{latex/charts/sampling}

\subsection{Model Details}
\label{subsec:models} 

We selected two open-source LLMs released before the 2024–25 NBA season to mitigate data contamination: \LlamaVer{3.1}{70B} (Jul.~2024; \citealt{grattafiori2024llama3herdmodels}) and \QwenVer{2.5-72B} (Sep.~2024; \citealt{Yang2024Qwen25TR}).  We accessed \LlamaVer{3.1}{70B} via the Together AI API \citep{togetherai2025}, and \QwenVer{2.5-72B} via Alibaba Cloud’s AI platform \citep{alibaba2025ai} using a registered account. We used identical decoding settings for both models. The model specifications are shown in \Cref{tab:model-specs}.

\paragraph{Prompt.}
We designed the prompt following best-practice guidance from Google \citep{google2025prompt} for zero-shot, task-focused generation: a clear role definition, step-by-step instructions, explicit output constraints, and style/tone specifications. We then customised the wording for each input (unstructured, row-structured, JSON) after multiple trials. The final prompt used in our experiments is shown in \Cref{fig:game_summary_prompt}.

\subsection{Annotation Protocol}
\label{subsec: annotation} 
   
We evaluate factual and semantic errors using a manual error-annotation protocol rather than automatic metrics such as ROUGE \citep{lin-2004-rouge} or BLEU \citep{papineni-etal-2002-bleu}. We refine the annotation scheme \citep{10.1016/j.csl.2023.101482} by splitting the original \emph{Word} category into \emph{Word-objective} (verifiable, fact-checkable wording errors) and \emph{Word-subjective} (opinion-based wording not grounded in the play-by-play). 

We use seven categories in total: \emph{Number}, \emph{Name}, \emph{Word-objective}, \emph{Word-subjective}, \emph{Context}, \emph{Not Checkable}, and \emph{Other}. These error categories are elaborated in \Cref{sec: error_category}. The manually annotated error counts form the basis for comparing performance across input structures in subsequent statistical analysis. 



\paragraph{Notation.}
For readability in examples and tables, we mark erroneous spans with coloured superscripts indicating the error category.
\begin{itemize}[leftmargin=1.2em, itemsep=0.2ex, topsep=0.2ex, parsep=0pt, partopsep=0pt]
  \item \markerrnumber{10}~$\to$~\emph{6} \hfill (U: Number)
  \item \markerrname{LeBron James}~$\to$~\emph{Anthony Davis} \hfill (N: Name)
  \item \markerrwordobj{free throw}~$\to$~\emph{layup} \hfill (Wo: Word-objective)
  \item \markerrwordsub{pivotal moment} \hfill (Ws: Word-subjective)
  \item \markerrcontext{Christian Wood}~$\to$~\emph{Cam Whitmore} \hfill (C: Context)
  \item \markerrnotcheckable{third straight victory} \hfill (X: Not Checkable)
  \item \markerrother{other} \hfill (O: Other)
\end{itemize}

\subsection{Hypotheses}
\label{sec: hypotheses} 

We test the following hypotheses about how input structure and model choice influence the number of factual and semantic error rates in LLM-generated summaries:

\begin{itemize}
    \item \textbf{H1 – Input Structure Effect:}  
    Error rates differ significantly across input formats; structured inputs (row-structured and JSON) are expected to produce fewer errors than unstructured input.
    \item \textbf{H2 – Model Effect:}  
    Error rates differ significantly between the two large language models (\LlamaVer{3.1}{70B} and \QwenVer{2.5-72B}), with one model expected to perform better overall.
    \item \textbf{H3 – Interaction Effect:}  
    The effect of input structure on error rates depends on the model, suggesting that some models perform better with certain input structures.
\end{itemize}

To test these hypotheses, we conduct a two-way repeated-measures ANOVA on the total number of annotated errors per 100 words in each summary. Input structure and model are treated as within-subjects factors.

\section{Results}
\label{sec: results} 

We manually annotated 3,312 errors across 180 play-by-play summaries generated for 30 NBA games (2 models $\times$ 3 input structures). \LlamaVer{3.1}{70B} averaged 408 words per summary, whereas \QwenVer{2.5-72B} averaged 505 words. To compare factual accuracy across both models and all three input structures, we report normalised error rates, i.e., errors per 100 words (see \Cref{tab:errors_overall}). The breakdown by error type is shown in \Cref{tab:errors_by_type}.

\input{latex/tables/error_categories}

\subsection{Analysis by Total Number of Errors}
\label{sec: totalerrors} 
\subsubsection{Distribution of Errors}

\Cref{fig:input-structure-summary} and \Cref{tab:descriptive_stats_summary} show the distribution of total annotated errors across 180 game summaries, grouped by models (\LlamaVer{3.1}{70B} and \QwenVer{2.5-72B}\footnote{Llama denotes \LlamaVer{3.1}{70B}; Qwen denotes \QwenVer{2.5-72B}. These model names refer exclusively to these versions throughout the paper.}) and input structures (Row, JSON and Unstructured). Both models produced fewer errors in structured inputs: JSON input resulted in the lowest errors (2.17 mean errors for Llama and 2.14 for Qwen), while the row structure had 3.20 for Llama and 2.92 for Qwen. Both models consistently recorded more errors with unstructured inputs (7.05 for Llama and 6.04 for Qwen).

\input{latex/tables/anova_two_way_repeated}

\input{latex/charts/descriptive_bar}

\paragraph{Two-way ANOVA Repeated-Measures.}

We conducted a two-way repeated-measures ANOVA to examine the effects of input structure and model (two factors as independent variables) on error rates. Each of the 30 games was evaluated under all six conditions (two models $\times$ three input structures), resulting in a within-subjects design. The dependent variable was the total number of errors normalised per 100 words.

As shown in \Cref{tab:anova_results}, the input structure had a highly significant effect on error rates ($p=1.79\times 10^{-27}$) with a large effect size (partial $\eta^2=0.813$), indicating that the way information was structured explains the majority of variance in summary errors. The model main effect was also significant ($p=3.60\times 10^{-4}$). The interaction between input structure and model was significant ($p=4.80\times 10^{-4}$). These findings highlight the dominant influence of input structure on error rates and motivate post hoc analyses to compare model–input structure combinations.

\subsubsection{Post hoc tests} We use Tukey's Honestly Significant Difference (HSD) test to compare the six model-input combinations as distinct conditions, since each produces a unique summary. Based on the results shown in \Cref{tab:tukey_combined}, we assess each hypothesis defined in \cref{sec: hypotheses}.

\paragraph{H1: Effect of Input Structure.}

In Tukey's HSD (\Cref{tab:tukey_combined}), the input structure significantly affects error rates across all pairwise comparisons (all \(p<0.05\)): JSON vs row (mean difference = 0.91), JSON vs unstructured (mean difference = 4.39), and row vs unstructured (mean difference = 3.48). We therefore conclude that input structure significantly influences error rates.

\paragraph{H2: Model Effect.}

Given the two-way ANOVA showed a significant model and input structure interaction in (\Cref{tab:anova_results}), we assess model differences within each input. Llama vs. Qwen differences are not significant after Tukey correction for JSON (\(p = 1.00\)) or row (\(p = 0.842\)), but there is a significant Llama vs. Qwen difference for unstructured (p = 0.0005).  Thus, we do not claim a uniform model advantage; model differences appear specifically in the unstructured input. The main effect of model should therefore be interpreted with caution.

\paragraph{H3: Effect of Input Structure and Model Interaction.}
Consistent with the significant interaction found in the ANOVA, within-model contrasts show a monotonic increase across input structures from JSON to row to unstructured.

\begin{itemize}

\item \textbf{JSON vs. Unstructured (within model):} As shown in \cref{tab:tukey_combined}, Llama's errors increase by 4.88 per 100 words when input changes from JSON to unstructured (\(p<0.05\)). Qwen shows a 3.90 increase for the same shift (\(p<0.05\)).

\item \textbf{JSON vs. Row and Row vs. Unstructured (within model):} Errors increase when moving from row to unstructured (\(p<0.05\)). From JSON to row, Llama shows a significant increase by 1.03 (\(p<0.05\)), and Qwen shows a significant increase by 0.78 (\(p<0.05\)).

\item \textbf{Same input across models (between models):} Some cross-model comparisons within the same input are not significant (e.g., Llama's JSON vs. Qwen's JSON: \(p=1.00\); Llama's row vs Qwen's row \(p=0.842\)), while unstructured differs \(p=0.0005\).

\end{itemize}

The results demonstrate that input structure significantly affects performance. Errors for both models increase from JSON to row to unstructured, but by different magnitudes; therefore, any model advantage is conditional on input (negligible for JSON and row, present for unstructured).

\input{latex/tables/input_chart}

\subsection{Analysis by Error Category}
\label{sec: error_category} 

As mentioned in \Cref{tab:errors_by_type}, we compute the error rate by normalising errors per hundred words for each category. \Cref{fig:error_and_input_type_comparison} presents a comparison of error rates for each category across three inputs. The bar chart also provides an overview of errors made by two models. Out of the seven error categories, we can observe that number, name, and word objective errors are predominant in all combinations. While word subjective and context are specific to certain models and input structures. 

In addition to the main hypotheses we tested in section 4.1, we conduct a two-way ANOVA with repeated measures for the five main error categories to understand the significance between these specific error types, models, and input structures.

\paragraph{Number errors.}
Numbers encode key information in basketball play-by-play data: it has point increments, cumulative scores, shot counts (two-pointer, three-pointer), shot distances (22ft, 2 ft), and period details in ordinals (first, second, third, and fourth). As shown in \Cref{fig:error_and_input_type_comparison}, both models exhibit their highest number-error rates on unstructured input, and lower rates on row and JSON. On the structured inputs, Llama commits higher number errors than Qwen. As reported in \Cref{tab:anova_number_errors}, a two-way repeated-measures ANOVA on number-error rates also shows significant model, input structure and model$\times$input interaction (all $p<0.05$)


\paragraph{Name errors.}
 Name errors occur when a model misnames players or teams in the summary. For example, swapping the winning team's name or attributing an action to the wrong player. Qwen makes more of these mistakes than Llama even on JSON input (see \Cref{fig:json_panel}). In \Cref{tab:anova_name_errors}, a two-way repeated-measures ANOVA shows significant main effects of model and input (both $p<0.05$), with no significant model$\times$input interaction.


\paragraph{Word-objective errors.}

 Word-objective errors are predominant in both models across all input structures. These errors occur when the model generates an incorrect description of actions in the game, e.g., `\textit{calling a turnover a rebound}' or claim a player `\textit{orchestrated a fast-break opportunity that led to a layup}' even when the team never lost possession. They also commit verb errors, such as `\textit{outscore}', `\textit{tied the game}' or `\textit{cut the deficit}' when the plays show no such changes. These mistakes highlight how even small wording errors can misrepresent the game moments. A two-way repeated-measures in \Cref{tab:anova_word_objective_errors} shows a significant input effect ($p<0.05$), with non-significant model and model$\times$input interaction.


\paragraph{Word-subjective errors.}
Word-subjective errors occur when a model adds opinion-based wording to an objective summary. Examples include calling a contest a `\textit{classic match},' praising a team for showing `flashes of brilliance' or `relentless efforts', describing a play as a `pivotal moment' or mentioning it a `memorable game in this season.' These phrases reflect personal judgement and depend on the annotator's interpretation rather than verifiable facts. They fall outside the factual error category, but we captured this error type to ensure that any subjective or opinion-based wording is flagged. Qwen generated more word subjective errors than Llama across all inputs (see \Cref{fig:error_and_input_type_comparison}). \Cref{tab:anova_word_subjective_errors} shows significant main effects of input and model ($p<0.05$), with no significant model$\times$input interaction. 



\paragraph{Context errors.}
Context errors occur when a summary leads readers to misinterpret the game, such as describing actions by players who did not participate in that period or game. We observed this error only in Qwen on unstructured inputs (\Cref{fig:error_and_input_type_comparison} and \Cref{fig:example_annotations}), where the model hallucinated player names when the input lacked complete and structured data (see \Cref{tab:anova_context_errors}).


\paragraph{Not checkable and Other errors.}

Not checkable errors occur when a summary states facts that cannot be verified from the game's per-period, box-score, or play-by-play data. Even if the information lies outside these sources (for example, `third straight victory'), we still flag it as \emph{Not checkable}. `Other' error type cover any mistakes that do not fit in the previous categories. Both of these errors occurred infrequently and showed no significant variation across input structures or models, so we did not perform further in-depth analysis on them.
\input{latex/tables/game_summaries_main}





\input{latex/tables/annotator_agreement}
\section{Discussion}
\label{sec: discussion} 

\Cref{fig:example_annotations} presents partial-game summaries with error annotations to demonstrate a few examples of factual errors produced by our models for three different inputs (taken from the first overtime of the game shown in \Cref{tab:row_example} and \Cref{tab:unstructured_ot1}).

\subsection{Input structure} 
We observe multiple factual errors in the partial game summary based on unstructured input (see \Cref{fig:main_example_annotations}). In this example, the Llama model struggles to produce a factual summary: it describes incorrect actions such as `a pair of three-pointers' and `a dunk and a layup,' uses wrong player names, and reports incorrect scores. Number errors are predominant; for instance, when the model generates `tie the game at 130–129,' it fails to reason correctly, as the actual score was 123.

\Cref{fig:example_annotations} shows that Qwen on \emph{unstructured input} struggles to generate the correct player names for the teams in the game. For example, it outputs Frank Ntilikina and Christian Wood, who were not on the Cavaliers or the Rockets, which constitutes a context error. One cause of these name and context errors is that the original play-by-play data (\Cref{subsec: preprocess}) represents players with initials and surnames (e.g.,`L. James' for LeBron James), which can lead the model to produce incorrect player names. More importantly, the unstructured data lacks atomic values: specific events and actions are not separated into meaningful columns or key-value pairs (see \Cref{tab:unstructured_ot1}).



The summaries from \emph{row-structured inputs} capture play-by-play data and events by clearly segregating entities (player names, team names), and action types (jump shots, layups, rebounds, distance shots, incremental points, cumulative scores) into meaningful column names with atomic values. This structure we hypothesise helps to preserve the information and reduce the factual errors to some extent. For example, the row-structured partial summary in \Cref{fig:main_example_annotations} correctly identifies several events (jump shot, three-pointer, tying the game) but still makes three word-objective errors, describing Young’s jump shot as a layup. Due to the challenging nature of over 450 plays in a game, even the hierarchical JSON struggles to render the right order of actions in some examples.


The summaries generated from \emph{hierarchical JSON} data (see \Cref{fig:play_json}) proved more factually accurate than the row-structure input. By encoding each event as a nested key-value pair, where time, team, play details, and score are separate fields, this clear separation preserves the exact relationship between the actions taken. Most names, numbers, and actions are correct (see \Cref{fig:main_example_annotations} and  \Cref{fig:example_annotations}), though James's `jump shot' is misrendered as a layup. 




\subsection{Models} 
In terms of the  summary style, Llama focuses on the first few plays and last few plays in each quarter. They also produce around 8 to 12 numerical figures to summarize one quarter, compared to Qwen’s 5–6 for the same period. For example, 

\begin{quote}
The Rockets responded with a \markerrnumber{10}-\markerrnumber{0} run, capped off by a dunk from Cam Whitmore, to take a 45-\markerrnumber{35} lead. \\
\markerrnumber{10}-\markerrnumber{0}: number errors, the correct runs during that time is 6-7 run. Llama makes these computation quite often and results in more number errors.\\
\markerrnumber{35}: number error, the opponent team score is 32.
\end{quote}

Qwen makes slightly more name mistakes than Llama (see \cref{fig:error_and_input_type_comparison} and \cref{fig:example_annotations}). We observed a few instances where both models showed bias towards star players in their summaries. For example,
\begin{quote}
The Lakers extended their lead with a series of baskets from \markerrname{LeBron James} and Anthony Davis. 
\end{quote}

In fact, the baskets came from Knecht and Davis, but both models incorrectly substituted LeBron James, reflecting a tendency, though not frequent, to favour well-known players in their summaries.

        
\textbf{Importance of prompt.} The instructions given in the prompt also play an important role in reducing a few errors in the summaries. After several trials, we applied the prompt mentioned in (\Cref{fig:game_summary_prompt}) for all our generations. Qwen's summaries adhered to the instructions (narrative style, structure, constraints), whereas Llama followed the instructions for narrative style and structure but struggled with constraints such as calculating individual player points (even when instructed not to) and maintaining the mandatory word limit of 450 words. The content focus differs for both models.

\subsection{Inter-annotator agreement} 
\label{subsec:iaa}

One of the authors manually annotated errors in all 180 generated summaries, and a second annotator independently annotated 9 summaries (5\% of the dataset) to assess inter-annotator agreement. We provided detailed guidelines describing error categories, examples, and instructions for marking errors. Annotators worked independently and marked errors at the token/phrase level.

Agreement was measured as exact-match overlap: for each category, we computed precision, recall, and unbiased F1 by alternately treating each annotator as the reference and averaging. The overall agreement was 80.3\% with unbiased F1 of 0.89, indicating strong agreement despite some variation across categories.

\section{Conclusion}
\label{sec: conclusion} 

Factual inaccuracies remain a major challenge for deploying LLMs in real-world applications. Our study shows that structured inputs, especially hierarchical JSON, substantially reduce errors in NBA play-by-play summaries: error rates dropped by up to 69\% for Llama and 65\% for Qwen compared to unstructured text. A two-way repeated-measures ANOVA confirmed that input structure explained over 80\% of the variance in error rates, with significant differences across all formats. As future work, we will develop an LLM-as-Judge framework to complement costly manual evaluation by automatically assessing atomic factual claims against play-by-play input logs. We will also extend this work to ice hockey play-by-play data to identify general data-structuring features that minimise factual inaccuracies across multiple sports datasets.

\section*{Limitations}

We acknowledge some limitations in this work. First, we only looked at NBA basketball game summaries. Second, we required annotators to have qualitative NBA knowledge to understand the play-by-play data. This requirement improved annotation accuracy but significantly reduced the pool of eligible annotators and limited scalability. Each summary took approximately 30 to 60 minutes to annotate, depending on its complexity, and one eligible annotator withdrew because of this time constraint. Third, we evaluated only two models (Llama and Qwen), which limits the generalizability of our findings to other large language models.


\section*{Ethics Statement}
We received approval from the University of Aberdeen Ethics Review Board to perform the inter-annotation experiment. This study is based entirely on publicly available NBA play-by-play data, which we preserved in its original structure without introducing additional bias. For error annotation, one author performed a full manual review of all 180 summaries, and a second annotator, who volunteered and provided informed consent, reviewed 9 summaries to help validate consistency. Annotators received detailed guidelines and example annotations, were free to withdraw at any time without penalty, and were compensated for their time as previously agreed upon before the annotation process. 

\section*{Acknowledgements}
We thank Javier González Corbelle for his hard work in helping with the annotations in this paper. We thank the anonymous reviewers for their detailed feedback and suggestions, which have significantly improved this work. We also thank the NLG (CLAN) reading group at the University of Aberdeen for their invaluable feedback.




\newpage
\appendix
\begin{appendices}
This appendix provides supplementary figures, tables, and prompts referenced in the main paper. We include the unstructured play-by-play input (\Cref{{tab:unstructured_ot1}}) for reference. The model hyperparameters and decoding settings used to generate the 180 game summaries are shown in \Cref{tab:model-specs}. We also report post-hoc Tukey HSD tests and a full set of ANOVA tables for each error type (number, name, word-objective, 
word-subjective, and context). These results complement the summary statistics in the main text 
by presenting full $p$-values and effect sizes. To further illustrate the annotation protocol, we provide extended annotated examples of 
game summaries with error labels (\Cref{fig:example_annotations}), as well as the full evaluation prompt template used for 
generation (\Cref{fig:game_summary_prompt}).  

\input{latex/tables/unstructured}

\input{latex/tables/descriptive_stats}
\label{sec:appendixanova}

\label{appendix a: anova_results}

\input{latex/tables/model_spec}
\input{latex/tables/post_hoc_tukey}
\input{latex/tables/number_anova}
\input{latex/tables/name_anova}
\input{latex/tables/word_objective_anova}
\input{latex/tables/word_subjective_anova}
\input{latex/tables/context_anova}

\input{latex/tables/example_annotations}


\input{latex/tables/prompt}

\end{appendices}
\end{document}

%% file: latex/tables/json_structure.tex
\begin{figure}[t!]
  \raggedright
  \small
  \begin{lstlisting}[language=json,breaklines=true]
{
  "time": {
    "period": "OT1",
    "clock": "2:13.0"
  },
  "team": "Los Angeles Lakers",
  "play_details": {
    "description": "LeBron James makes 3-pt jump shot from 25 ft",
    "event": {
      "type": "Shot",
      "action": "Jump Shot",
      "outcome": "Made",
      "distance": "25 ft",
      "points": 3
    },
    "players": {
      "primary_player": "LeBron James"
    }
  },
  "score": {
    "points_scored": "LAL: 3",
    "cumulative": {
      "LAL": 127,
      "ATL": 125
    }
  }
}
  \end{lstlisting}

  \caption{\textbf{Hierarchical JSON representation of a single play-by-play event.}  
\textit{(i) Event:} LeBron James makes a three-point jump shot from 25~ft; 
\textit{(ii) Time:} period (first overtime) and game clock (i.e., time remaining in that period); \textit{(iii) Score:} points scored from the play and the teams' cumulative scores.}

  \label{fig:play_json}
\end{figure}

%% file: latex/tables/row_structure.tex
\begin{table*}[t]
\centering
\small
\begin{subtable}[t]{\linewidth}
\centering
\resizebox{\linewidth}{!}{%
\begin{tabular}{@{} l l l l l p{5cm} @{}} 
\toprule
\textbf{Time} & \textbf{Period} & \textbf{Team} & \textbf{Primary Player} & \textbf{Secondary Player} & \textbf{Play Description} \\
\midrule

2:56.0 & OT1 & Los Angeles Lakers & Anthony Davis & N/A & Anthony Davis makes free throw 1 of 2 \\
2:56.0 & OT1 & Los Angeles Lakers & Anthony Davis & N/A & Anthony Davis makes free throw 2 of 2 \\
2:45.0 & OT1 & Atlanta Hawks & De'Andre Hunter & LeBron James & De'Andre Hunter misses 2-pt layup from 10 ft (block by LeBron James) \\
2:44.0 & OT1 & Atlanta Hawks & N/A & N/A & Offensive rebound by Team \\
2:33.0 & OT1 & Atlanta Hawks & Trae Young & N/A & Trae Young makes 2-pt layup from 2 ft \\
2:13.0 & OT1 & Los Angeles Lakers & LeBron James & N/A & LeBron James makes 3-pt jump shot from 25 ft \\

\bottomrule
\end{tabular}
}
\subcaption{Temporal information, Team and Player details, and Play description in row-structured input.}
\end{subtable}

\vspace{1em}

\begin{subtable}[t]{\linewidth}
\centering
\resizebox{\linewidth}{!}{%
\begin{tabular}{@{} l l l l l l l @{}}  
\toprule
\textbf{Event Type} & \textbf{Action Type} & \textbf{Outcome} & \textbf{Distance (ft)} & \textbf{Current Points Scored} & \textbf{LAL Cum. Score} & \textbf{ATL Cum. Score} \\
\midrule
Shot & Free Throw & Made & N/A & LAL: 1 & LAL: 123 & ATL: 123 \\
Shot & Free Throw & Made & N/A & LAL: 1 & LAL: 124 & ATL: 123 \\
Shot & Layup & Missed & 10 & No points scored & LAL: 124 & ATL: 123 \\
Rebound & Offensive & N/A & N/A & No points scored & LAL: 124 & ATL: 123 \\
Shot & Layup & Made & 2 & ATL: 2 & LAL: 124 & ATL: 125 \\
Shot & Jump Shot & Made & 25 & LAL: 3 & LAL: 127 & ATL: 125 \\
\bottomrule
\end{tabular}
}
\subcaption{Structured event and score details in row-structured input.}
\end{subtable}

\caption{\textbf{Excerpt of row-structured play-by-play input representation (partial game).} This is a \emph{single table} split into upper and lower panels for readability. `N/A' indicates that the attribute does not apply. In the released dataset, such fields are stored as \texttt{None}.}
\label{tab:row_example}
\end{table*}

%% file: latex/charts/sampling.tex
\begin{figure}[t]
    \centering
    \definecolor{colorblindlightgreen}{rgb}{0.67, 0.9, 0.65}
    \definecolor{dimgray85}{RGB}{85,85,85}
    \definecolor{gainsboro229}{RGB}{229,229,229}
    \definecolor{lightsalmon255159159}{RGB}{255,159,159}
    \figline[1.0pt]
    \begin{tikzpicture}
        \begin{axis}[
          ybar,
          bar width=12pt,
          width=0.75\linewidth,
          height=0.5\linewidth,
          enlarge x limits=0.20,
          ymin=0,
          axis lines*=left,
          axis line style={black},
          tick pos=left,                 
          grid=none,                          
          xmajorgrids=false, ymajorgrids=false, 
          x grid style={draw=none}, y grid style={draw=none}, 
          xlabel={Absolute score difference between two teams (points)},
          ylabel={No.\ of games},
          symbolic x coords={{$\leq$3},{4--9},{10--19},{$\geq$20}},
          xtick=data,
          tick label style={font=\small},
          label style={font=\small},
          nodes near coords,
          nodes near coords align={vertical},
          every node near coord/.append style={
            font=\footnotesize,
            /pgf/number format/precision=0,
            text=black
          },
        ]
          \addplot+[fill=lightsalmon255159159, draw=none] coordinates {
            ({$\leq$3},55) (4--9,130) (10--19,136) ({$\geq$20},98)
          };
    
          \addplot+[fill=dimgray85, draw=none] coordinates {
            ({$\leq$3},4) (4--9,9) (10--19,10) ({$\geq$20},7)
          };
        \end{axis}
    \end{tikzpicture}
    
    \caption{\textbf{Games by final margin.} For each stratum (\emph{absolute score difference: $\leq 3$, 4--9, 10--19, $\geq 20$}), the left bar shows the total number of NBA games in Dec.~2024 -- Jan.~2025, and the right bar shows the stratified sample used in our study. Values above bars are counts.}
    \label{fig:strata_doublebar}
    \figline[1.0pt]
\end{figure}
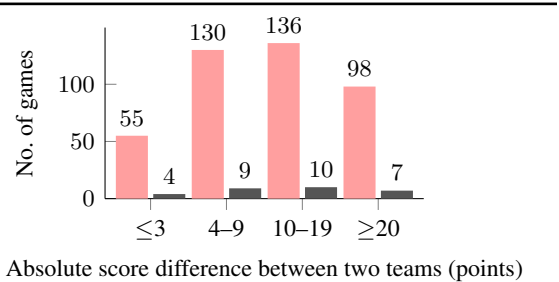

%% file: latex/tables/error_categories.tex
\begin{table}[ht]
  \centering
  \begin{adjustbox}{max width=\columnwidth}
    \begin{tabular}{l c c}
      \toprule
      \textbf{Metric} & \textbf{\LlamaVer{3.1}{70B}} & \textbf{\QwenVer{2.5-72B}} \\
      \midrule
      Total errors & 1{,}575 & 1{,}737 \\
      Total words  & 36{,}709 & 45{,}474 \\
      \textbf{Error rate} (per 100 words) & \textbf{4.29} & \textbf{3.82} \\
      \bottomrule
    \end{tabular}
  \end{adjustbox}
  \caption{\textbf{Total errors and normalised error rates} aggregated across all 30 summaries and three input formats. Normalised error rate $=$ (total errors / total words) $\times 100$.}
  \label{tab:errors_overall}
\end{table}

\begin{table}[ht]
  \centering
  \begin{adjustbox}{max width=\columnwidth}
    \begin{tabular}{l c c}
      \toprule
      \textbf{Error type} & \textbf{\LlamaVer{3.1}{70B}} & \textbf{\QwenVer{2.5-72B}} \\
      \midrule
      \markerrnumber{Number}            & 1.82 & 0.93 \\
      \markerrname{Name}                & 0.66 & 0.80 \\
      \markerrwordobj{Word-objective}   & 1.35 & 1.25 \\
      \markerrwordsub{Word-subjective}  & 0.31 & 0.61 \\
      \markerrcontext{Context}          & 0.10 & 0.19 \\
      \markerrnotcheckable{Not checkable} & 0.05 & 0.04 \\
      \markerrother{Other}              & 0.00 & 0.00 \\
      \bottomrule
    \end{tabular}
  \end{adjustbox}
  \caption{\textbf{Normalised error rates by error type and model}, aggregated across all summaries and three input formats. Each rate is computed as (errors of that type / total words) $\times 100$. The sum of all error types equals the overall error rate reported in \Cref{tab:errors_overall}, up to rounding.}
  \label{tab:errors_by_type}
\end{table}

%% file: latex/tables/anova_two_way_repeated.tex
\begin{table}[!t]
    \centering
    \small

    \begin{tabular}{lcc}
        \toprule
        \textbf{Source} & \textbf{$p$ (GG)} & \textbf{partial $\eta^2$} \\
        \midrule
        Input Structure & $1.79\times10^{-27}$ & 0.813 \\
        Model & $3.60\times10^{-4}$  & 0.056 \\
        Input Structure $\times$ Model & $4.80\times10^{-4}$  & 0.050 \\
        \bottomrule
    \end{tabular}
\caption{\textbf{Two‐way repeated measures ANOVA.} Greenhouse–Geisser corrected $p$‐values (all p < 0.05; statistically significant). We also report effect size, partial $\eta^2$ for main effects and their interaction. Post hoc Tukey HSD results are in \Cref{tab:tukey_combined}.}
\label{tab:anova_results}
\end{table}

%% file: latex/charts/descriptive_bar.tex
\definecolor{llamacol}{RGB}{142,192,222} 
\definecolor{qwencol}{RGB}{220,200,180} 

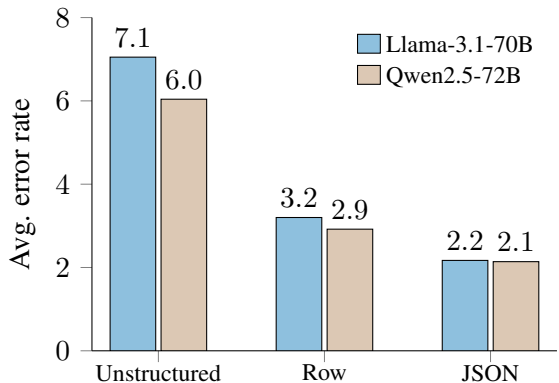
\begin{figure}[t]
\centering
\definecolor{green}{RGB}{142,192,222}
\definecolor{beige}{RGB}{220,200,180}

\begin{tikzpicture}
  \begin{axis}[
    ybar,
    bar width=0.6cm,
    width=1.0\linewidth,
    height=6cm,
    enlarge x limits=0.20,
    ymin=0, ymax=8.0,
    ytick={0,2,...,8},
    symbolic x coords={Unstructured, Row, JSON},
    xtick=data,
    xticklabel style={
      rotate=0,
      anchor=center,
      font=\small,
      yshift=-4pt
    },
    ylabel={Avg. error rate},
    axis lines*=left,
    axis line style={black},
    xtick pos=bottom, ytick pos=left,
    tick align=outside,
    grid=none,
    legend style={
      at={(0.98,0.98)},
      anchor=north east,
      cells={anchor=west},
      draw=none,
      fill=none,
      font=\small
    },
    legend image code/.code={
      \draw[#1, fill=#1] (0cm,-0.1cm) rectangle (0.3cm,0.15cm);
    },
    nodes near coords,
    every node near coord/.append style={
      /pgf/number format/.cd,
      fixed,
      fixed zerofill,
      precision=1
    },
  ]

    \addplot[fill=green] coordinates {
      (Unstructured, 7.05)
      (Row,           3.20)
      (JSON,          2.17)
    };
    \addplot[fill=beige] coordinates {
      (Unstructured, 6.04)
      (Row,           2.92)
      (JSON,          2.14)
    };
    \legend{Llama-3.1-70B, Qwen2.5-72B}
  \end{axis}
\end{tikzpicture}

\caption{\textbf{Average error rates by input structure for two models (errors per 100 words).}}
\label{fig:input-structure-summary}
\end{figure}

%% file: latex/tables/input_chart.tex
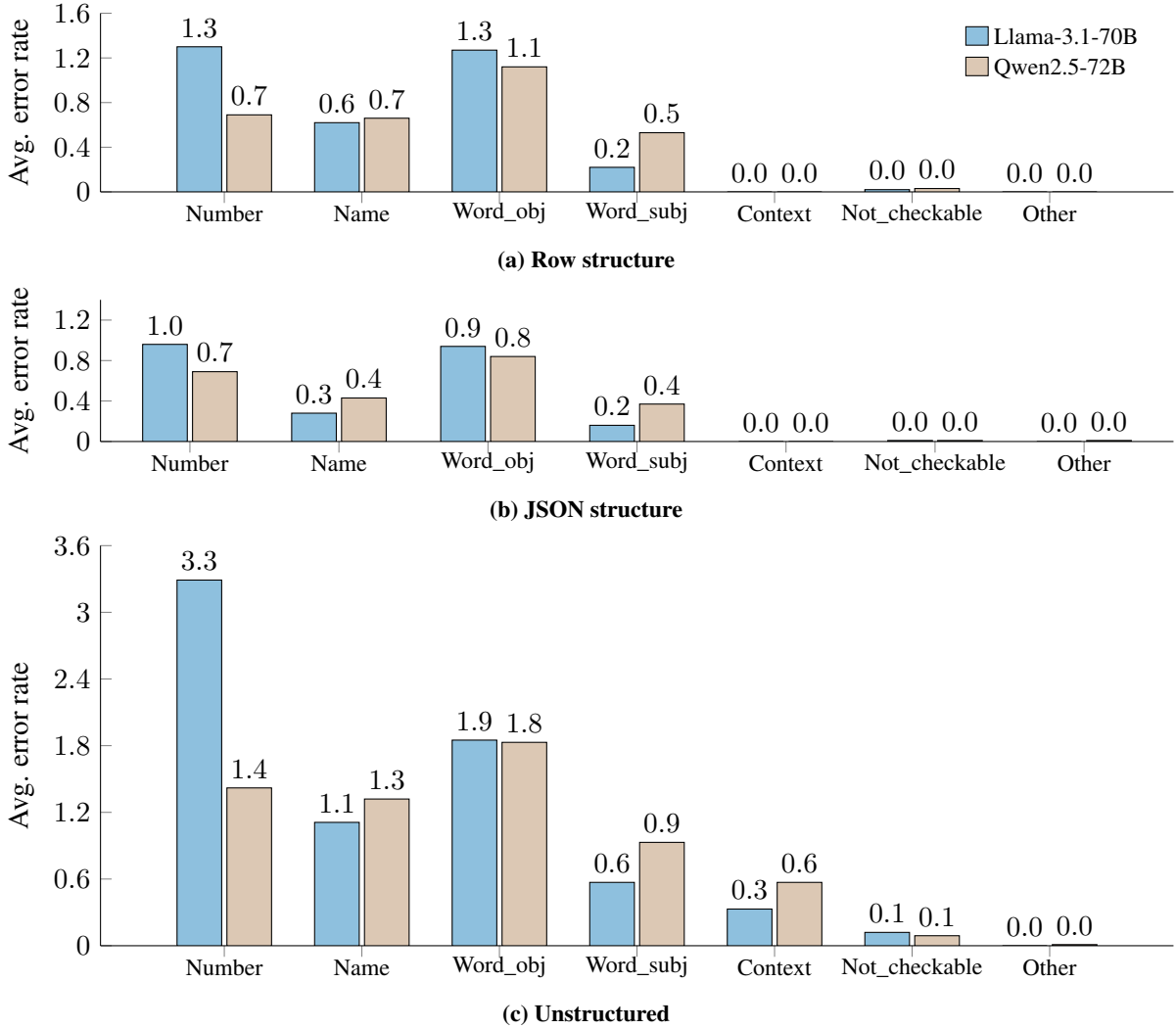
\begin{figure*}[t]
  \centering
  \definecolor{green}{RGB}{142,192,222}
  \definecolor{beige}{RGB}{220,200,180}

  \begin{subfigure}{\textwidth}
    \centering
    \begin{tikzpicture}
      \begin{axis}[
        ybar,
        bar width=0.6cm,
        width=\linewidth,
        height=4cm,
        enlarge x limits=0.15,
        ymin=0, ymax=1.6,
        ytick={0,0.4,...,1.6},
        symbolic x coords={Number, Name, Word\_obj, Word\_subj, Context, Not\_checkable, Other},
        xtick=data,
        xticklabel style={rotate=0, anchor=center, font=\small, yshift=-4pt},
        ylabel={Avg. error rate},
        axis lines*=left,
        axis line style={black},
        xtick pos=bottom, ytick pos=left,
        nodes near coords,
        every node near coord/.append style={/pgf/number format/.cd, fixed, fixed zerofill, precision=1},
        legend style={at={(0.98,0.98)}, anchor=north east, cells={anchor=west}, draw=none, fill=none, font=\small},
        legend image code/.code={\draw[#1, fill=#1] (0cm,-0.1cm) rectangle (0.3cm,0.15cm);},
      ]
        \addplot[fill=green] coordinates {
          (Number,1.30) (Name,0.62) (Word\_obj,1.27) (Word\_subj,0.22)
          (Context,0.00) (Not\_checkable,0.02) (Other,0.00)
        };
        \addplot[fill=beige] coordinates {
          (Number,0.69) (Name,0.66) (Word\_obj,1.12) (Word\_subj,0.53)
          (Context,0.00) (Not\_checkable,0.03) (Other,0.00)
        };
        \legend{Llama-3.1-70B, Qwen2.5-72B}
      \end{axis}
    \end{tikzpicture}
    \subcaption{\textbf{Row structure}}
    \label{fig:row_panel}
  \end{subfigure}\hfill

  \begin{subfigure}{\textwidth}
    \centering
    \begin{tikzpicture}
      \begin{axis}[
        ybar,
        bar width=0.6cm,
        width=\linewidth,
        height=3.5cm,
        enlarge x limits=0.1,
        ymin=0, ymax=1.4,
        ytick={0,0.4,...,1.6},
        symbolic x coords={Number, Name, Word\_obj, Word\_subj, Context, Not\_checkable, Other},
        xtick=data,
        xticklabel style={rotate=0, anchor=center, font=\small, yshift=-4pt},
        ylabel={Avg. error rate},
        axis lines*=left,
        axis line style={black},
        xtick pos=bottom, ytick pos=left,
        nodes near coords,
        every node near coord/.append style={/pgf/number format/.cd, fixed, fixed zerofill, precision=1},
      ]
        \addplot[fill=green] coordinates {
          (Number,0.96) (Name,0.28) (Word\_obj,0.94) (Word\_subj,0.16)
          (Context,0.00) (Not\_checkable,0.01) (Other,0.00)
        };
        \addplot[fill=beige] coordinates {
          (Number,0.69) (Name,0.43) (Word\_obj,0.84) (Word\_subj,0.37)
          (Context,0.00) (Not\_checkable,0.01) (Other,0.01)
        };
      \end{axis}
    \end{tikzpicture}
    \subcaption{\textbf{JSON structure}}
    \label{fig:json_panel}
  \end{subfigure}\hfill

  \begin{subfigure}{\textwidth}
    \centering
    \begin{tikzpicture}
      \begin{axis}[
        ybar,
        bar width=0.6cm,
        width=\linewidth,
        height=7cm,
        ymin=0, ymax=3.6,
        ytick={0,0.6,...,3.6},
        enlarge x limits=0.15,
        symbolic x coords={Number, Name, Word\_obj, Word\_subj, Context, Not\_checkable, Other},
        xtick=data,
        xticklabel style={rotate=0, anchor=center, font=\small, yshift=-4pt},
        ylabel={Avg. error rate},
        axis lines*=left,
        axis line style={black},
        xtick pos=bottom, ytick pos=left,
        nodes near coords,
        every node near coord/.append style={/pgf/number format/.cd, fixed, fixed zerofill, precision=1},
      ]
        \addplot[fill=green] coordinates {
          (Number,3.29) (Name,1.11) (Word\_obj,1.85) (Word\_subj,0.57)
          (Context,0.33) (Not\_checkable,0.12) (Other,0.00)
        };
        \addplot[fill=beige] coordinates {
          (Number,1.42) (Name,1.32) (Word\_obj,1.83) (Word\_subj,0.93)
          (Context,0.57) (Not\_checkable,0.09) (Other,0.01)
        };
      \end{axis}
    \end{tikzpicture}
    \subcaption{\textbf{Unstructured}}
    \label{fig:unstruct_panel}
  \end{subfigure}

  \caption{\textbf{Average error rates (errors per 100 words) by error type and input structure.}
  Each panel reports \textit{average error rates across the 30 summaries} for that input format (Row, JSON, Unstructured), comparing \LlamaVer{3.1}{70B} and \QwenVer{2.5-72B} across seven error categories. Structured inputs (Row and JSON) yield significantly fewer total errors than unstructured inputs, with JSON providing the largest reduction.}
  \label{fig:error_and_input_type_comparison}
\end{figure*}

%% file: latex/tables/game_summaries_main.tex
\begin{figure*}[tbp]
{\fontsize{10}{11}\selectfont
\setlength{\emergencystretch}{2em}
    \lineacross{}
    \begin{center}
    \textbf{Partial-game summary from Llama-Unstructured input}
  \end{center}

     \vspace{2mm}

     The Lakers fought back, however, with James and Davis leading the charge. James hit \markerrwordobj{a pair of three-pointers}, and Davis added \markerrwordobj{a dunk and a layup} to tie the game at \markerrnumber{130}-\markerrnumber{129}. 
     
     Throughout the game, James and Davis were the Lakers' top performers, combining for \markerrnumber{63} points. Young and \markerrname{Bogdanovic} led the Hawks, combining for \markerrnumber{61}points.
     
   \textbf{ List of errors.}
    \vspace{-3mm}
    \begin{itemize}[leftmargin=*]
    \item \markerrwordobj{a pair of three-pointers}: incorrect word objective phrases from events, correct event/phrase is `a three-pointer'.
    \item \markerrwordobj{a dunk and a layup}: incorrect events (word objective), the player made a `free-throw'.
    \item \markerrnumber{130} - \markerrnumber{129}: incorrect scores (numbers), the correct scores are `123-123.'
    \item \markerrnumber{63}: incorrect score (number), the correct number/score is `77.' 
    \item \markerrname{Bogdanovic}: incorrect player name, the correct player name is `Hunter.' 
    \item \markerrnumber{61}: incorrect score (number), the correct number/score is `75.'
    \end{itemize}
    
    \begin{center}
    \textbf{Partial-game summary from Qwen-Row input} 
    \end{center}
    The game saw several lead changes, with Anthony Davis making \markerrwordobj{a pair of free throws} to tie the game at 123. LeBron James then hit a three-pointer to give the Lakers a 127-125 lead, but Trae Young answered \markerrwordobj{with a layup} to tie it again. 
    
    The Lakers regained the lead with a three-pointer from Gabe Vincent, but the Hawks stayed close with a jump shot from Dyson Daniels. LeBron James finished with a \markerrwordobj{double}-double, contributing significantly on both ends of the court.   
    
   \textbf{ List of errors.}
    \vspace{-3mm}
    \begin{itemize}[leftmargin=*]
    \item \markerrwordobj{a pair of free throws}: the game was tied at 123 when Davis `made a free-throw.' 
    \item \markerrwordobj{with a layup}: incorrect events (word objective), the player made `a jump shot.'
    \item \markerrwordobj{double}: incorrect word (objective), LeBron James finished with a `triple-double.'
 
    \end{itemize}

    \begin{center}
    \textbf{Partial-game summary from Qwen-JSON input} 
    \end{center}
    The Hawks took an early lead with a layup from Onyeka Okongwu, but the Lakers quickly \markerrwordobj{tied it} with a three-pointer from Max Christie. 
    
    Throughout the game, LeBron James and Anthony Davis were the standout players for the Lakers, contributing significantly with scoring, rebounds, and assists.

    \textbf{ List of errors.}
    \vspace{-3mm}
    \begin{itemize}[leftmargin=*]
    \item \markerrwordobj{tied it}: word objective error, the Lakers `responded with' a three-pointer, they did not tie the game.

    \end{itemize}

    \begin{center}
    \textbf{Partial-game summary from Llama-JSON input} 
    \end{center}
    In overtime, the Hawks took the lead early, with Onyeka Okongwu making a layup to put them up 121-119. The Lakers fought back, with James making a 3-pointer to put them up 127-125. The Hawks regained the lead, with Trae Young making a 3-pointer to put them up 134-132 with 8 seconds left. The Lakers were unable to score on their final possession, and the Hawks emerged victorious.

    \textbf{ List of errors.} 
       \vspace{-3mm}
    \begin{itemize}[leftmargin=*]
        \item No errors in this partial summary.
    \end{itemize}
 
       \vspace{-3mm}

    \caption{\textbf{Examples of partial game summaries with error annotations, illustrating errors from the same game and period under the three input types, Row, JSON, and Unstructured} (see \Cref{tab:row_example}). Corrections are included for clarity.  \emph{Notation:} superscripts N, U, WO, WS, C, O, and X indicate the error categories: Name, Number, Word-objective, Word-subjective, Context, Other, and Not-checkable; this is used here for readability; see the Annotation Protocol \Cref{subsec: annotation} for definitions.}

    \lineacross{}
}
    \label{fig:main_example_annotations}
\end{figure*}

%% file: latex/tables/annotator_agreement.tex
\begin{table*}[!t]
\centering
\small
\begin{tabular}{p{2.2cm} c c c c c c c c}
\toprule
\textbf{Category} &
\makecell{\textbf{Total}\\\textbf{errors}} &
\makecell{\textbf{Both}\\\textbf{agree}} &
\makecell{\textbf{A1}\\\textbf{only}} &
\makecell{\textbf{A2}\\\textbf{only}} &
\makecell{\textbf{Agreement}\\\textbf{(\%)}} &
\makecell{\textbf{Precision}\\\textbf{(\%)}} &
\makecell{\textbf{Recall}\\\textbf{(\%)}} &
\makecell{\textbf{Unbiased}\\\textbf{F$_1$ (\%)}} \\
\midrule
Number & 112 & 108 & 0 & 4  & 96.4 & 100.0 &  98.2 &  98.2 \\
Name &  21 &  15 & 1 & 5 & 71.4 & 93.8 & 84.4 &  84.4 \\
Word\_objective & 23 & 13 & 4 & 6 & 56.5 & 76.5 & 72.4 & 72.4 \\
Word\_subjective & 13 & 4 & 2 & 7 & 30.7 &  66.7 &  51.5 & 51.5 \\
Context & 9 & 6 & 1 & 2 & 66.6 & 85.7 & 80.4 &  80.4 \\
Not\_checkable & 8 & 5 & 0 & 3 & 62.5 & 100.0 &  81.2 &  81.2 \\
Other & 2 & 0 & 2 & 0 & 0.0 & 0.0 & N/A & N/A \\
\midrule
\textbf{Overall} & 188 & 151 & 10 & 27 & \textbf{80.3} & \textbf{93.8} & \textbf{89.3} & \textbf{89.3} \\
\bottomrule
\end{tabular}
\caption{\textbf{Inter‑Annotator Agreement by Error Category.} Agreement is highest for \emph{Number} errors and lowest for \emph{Word-subjective} errors.} 
\label{tab:iaa_extended}

\end{table*}


%% file: latex/tables/unstructured.tex
\begin{table*}[ht]
\centering
\small
\resizebox{\linewidth}{!}{%
\begin{tabularx}{\textwidth}{l X c c c X}
\toprule
\textbf{Time} & \textbf{LA Lakers} & \na\ & \textbf{Score} & \na\ & \textbf{Atlanta} \\
\midrule
5:00.0 & Start of 1st overtime & \na\ & \na\ & \na\ & \na\ \\
5:00.0 & Jump ball: A. Davis vs.~O. Okongwu (J. Johnson gains possession) & \na\ & \na\ & \na\ & \na\ \\
3:05.0 & \na\ & \na\ & 122--123 & \na\ & Turnover by J. Johnson (bad pass; steal by G. Vincent) \\
2:56.0 & Shooting foul by O. Okongwu (drawn by A. Davis) & \na\ & 122--123 & \na\ & \na\ \\
2:56.0 & \na\ & \na\ & 122--123 & \na\ & C. Capela enters the game for O. Okongwu \\
2:56.0 & A. Davis makes free throw 1 of 2 & 1.0 & 123--123 & \na\ & \na\ \\
2:56.0 & A. Davis makes free throw 2 of 2 & 1.0 & 124--123 & \na\ & \na\ \\
2:45.0 & \na\ & \na\ & 124--123 & \na\ & D. Hunter misses 2-pt layup from 10 ft (block by L. James) \\
2:44.0 & \na\ & \na\ & 124--123 & \na\ & Offensive rebound by Team \\
2:33.0 & \na\ & \na\ & 124--125 & 2.0 & T. Young makes 2-pt layup from 2 ft \\
2:13.0 & L. James makes 3-pt jump shot from 25 ft & 3.0 & 127--125 & \na\ & \na\ \\
\bottomrule
\end{tabularx}
}
\caption[Unstructured play-by-play (OT1).]{\textbf{Excerpt of unstructured play-by-play input (partial game).} Raw log text from \href{https://www.basketball-reference.com/boxscores/pbp/202412060ATL.html\#q5}{Basketball-Reference (Overtime)}. No typed schema in the header; event semantics are embedded in free-text team columns, and cells are \emph{non-atomic}; one span may combine event type, participants, outcome, distance, and points. 
Fields that do not apply to a given event are shown as \na\ for reference. Columns show team descriptions, per-event points, and the cumulative score.}

\label{tab:unstructured_ot1}
\end{table*}

%% file: latex/tables/descriptive_stats.tex
\begin{table*}[t]
\centering
\begin{adjustbox}{max width=\textwidth}
\begin{tabular}{llcccccccccc}
\toprule
\textbf{Model} & \textbf{Input format} & \textbf{Mean} & \textbf{Median} & \textbf{Mode} & \textbf{Std Dev} & \textbf{Range} & \textbf{Min} & \textbf{Max} & \textbf{Skew} & \textbf{Kurt.} & \textbf{95\% CI} \\
\midrule
Llama & Row & 3.20 & 2.99 & 2.02 & 0.98 & 4.15 & 1.61 & 5.76 & 0.63 & 0.24  & [2.84, 3.57] \\
Llama & JSON & 2.17 & 2.02 & 1.32 & 0.64 & 2.32 & 1.32 & 3.64 & 0.55 & -0.64 & [1.93, 2.41] \\
Llama & Unstructured & 7.05 & 6.83 & 5.32 & 1.44 & 6.56 & 5.32 & 11.88 & 1.69 & 3.55 & [6.51, 7.59] \\
Qwen  & Row & 2.92 & 2.74 & 2.46 & 0.58 & 2.09 & 2.24 & 4.33 & 1.12 & 0.36 & [2.71, 3.14] \\
Qwen  & JSON & 2.14 & 2.00 & 1.06 & 0.76 & 3.03 & 1.06 & 4.09 & 0.84 & 0.22 & [1.85, 2.42] \\
Qwen  & Unstructured & 6.04 & 5.79 & 5.17 & 0.87 & 3.38 & 4.95 & 8.33 & 0.95 & 0.14 & [5.71, 6.37] \\
\bottomrule
\end{tabular}
\end{adjustbox}
\caption{\textbf{Descriptive statistics of normalised error rates across two models and three input structures.} The table reports measures of central tendency (mean, median, mode), variability (standard deviation, range, minimum, maximum), skewness, kurtosis, and 95\% confidence intervals.}
\label{tab:descriptive_stats_summary}
\end{table*}

%% file: latex/tables/model_spec.tex
\begin{table*}[t]
\centering
\small
\begin{adjustbox}{max width=\textwidth}
\begin{tabular}{l c c l c c c l}
\toprule
\textbf{Model} & \textbf{Parameters} & \textbf{Release} & \textbf{Prompt} &
\textbf{Temperature} & \textbf{Top-$p$} & \textbf{Top-$k$} & \textbf{API Platform} \\
\midrule
\LlamaVer{3.1}{70B} & 70B & Jul.~2024 & Zero-shot & 0 & 0.95 & 50 & Together AI API \\
\QwenVer{2.5-72B}   & 72B & Sep.~2024 & Zero-shot & 0 & 0.95 & 50 & Qwen (Alibaba) \\
\bottomrule
\end{tabular}
\end{adjustbox}
\caption{\textbf{Model specification and decoding settings used in all runs.}}
\label{tab:model-specs}
\end{table*}

%% file: latex/tables/post_hoc_tukey.tex
\begin{table*}[htbp]
\small
\centering
\begin{adjustbox}{max width=\textwidth}
\begin{tabular}{llccccc}
\toprule
\textbf{Group 1} & \textbf{Group 2} & \textbf{Mean Diff} & \textbf{$p$ (Tukey-adjusted)} & \textbf{Lower} & \textbf{Upper} & \textbf{Significant} \\
\midrule
\multicolumn{7}{l}{\textbf{(a) Input Structure}} \\
JSON  & Row          & 0.9102 & 0.000 & 0.4938 & 1.3265 & \textbf{Yes} \\
JSON  & Unstructured & 4.3913 & 0.000 & 3.9750 & 4.8077 & \textbf{Yes} \\
Row   & Unstructured & 3.4812 & 0.000 & 3.0648 & 3.8975 & \textbf{Yes} \\
\midrule
\multicolumn{7}{l}{\textbf{(b) Model}} \\
Llama & Qwen        & -0.4412 & 0.165 & -1.0651 & 0.1827 & No \\
\midrule
\multicolumn{7}{l}{\textbf{(c) Interaction: Input Structure × Model}} \\
Llama\_JSON & Llama\_Row & 1.0360 & 0.0003 & 0.3492 & 1.7228 & \textbf{Yes} \\
Llama\_JSON & Llama\_Unstructured & 4.8803 & 0.0000 & 4.1935 & 5.5671 & \textbf{Yes} \\
Llama\_JSON & Qwen\_JSON & -0.0313 & 1.0000 & -0.7181 & 0.6555 & No \\
Llama\_JSON & Qwen\_Row & 0.7530 & 0.0226 & 0.0662 & 1.4398 & \textbf{Yes} \\
Llama\_JSON & Qwen\_Unstructured  & 3.8710 & 0.0000 & 3.1842 & 4.5578 & \textbf{Yes} \\
Llama\_Row & Llama\_Unstructured & 3.8443 & 0.0000 & 3.1575 & 4.5311 & \textbf{Yes} \\
Llama\_Row & Qwen\_JSON & -1.0673 & 0.0002 & -1.7541 & -0.3805 & \textbf{Yes} \\
Llama\_Row & Qwen\_Row & -0.2830 & 0.8424 & -0.9698 & 0.4038 & No \\
Llama\_Row & Qwen\_Unstructured  & 2.8350 & 0.0000 & 2.1482 & 3.5218 & \textbf{Yes} \\
Llama\_Unstructured & Qwen\_JSON & -4.9117 & 0.0000 & -5.5985 & -4.2249 & \textbf{Yes} \\
Llama\_Unstructured & Qwen\_Row & -4.1273 & 0.0000 & -4.8141 & -3.4405 & \textbf{Yes} \\
Llama\_Unstructured & Qwen\_Unstructured & -1.0093 & 0.0005 & -1.6961 & -0.3225 & \textbf{Yes} \\
Qwen\_JSON & Qwen\_Row & 0.7843 & 0.0151 & 0.0975 & 1.4711 & \textbf{Yes} \\
Qwen\_JSON & Qwen\_Unstructured  & 3.9023 & 0.0000 & 3.2155 & 4.5891 & \textbf{Yes} \\
Qwen\_Row & Qwen\_Unstructured  & 3.1180 & 0.0000 & 2.4312 & 3.8048 & \textbf{Yes} \\
\bottomrule
\end{tabular}
\end{adjustbox}
\vspace{2px}
\caption{\textbf{Tukey HSD post hoc test results.} The table reports pairwise differences in mean error rates between groups, with Tukey-adjusted $p$-values ($p$-adj), 95\% confidence intervals (Lower, Upper), and a significance flag (Significant). Mean Diff is computed as Group 1 minus Group 2, so positive values indicate Group 1 has a higher mean than Group 2. \textbf{Section (a)} compares input structure conditions (Row, JSON, and Unstructured) and shows significant pairwise differences. \textbf{Section (b)} compares Models (Llama vs. Qwen). This contrast is not significant, indicating no reliable difference in overall error rates. \textbf{Section (c)} shows Interaction contrasts between specific Model and Input Structure combinations (e.g., Llama\_JSON vs. Llama\_Row). Several significant interactions exist between Model and Input Structure combinations.}
\label{tab:tukey_combined}
\vspace{4px}
\end{table*}

%% file: latex/tables/number_anova.tex
\begin{table*}[htbp]  
\centering  
\begin{adjustbox}{max width=\textwidth}
\begin{tabular}{lcccccc}  
\hline  
\textbf{Source} & \textbf{SS} & \textbf{DF} & \textbf{MS} & \textbf{F} & \textbf{$p$-unc} & $\eta_p^2$ \\  
\hline  
Input Structure            & 84.24  & 2   & 42.12  & 57.55  & $6.54\times10^{-20}$  & 0.398 \\  
Model                      & 37.90  & 1   & 37.90  & 51.80  & $1.77\times10^{-11}$  & 0.229 \\  

Model $\times$ Input Structure & 21.32  & 2   & 10.66  & 14.57  & $1.41\times10^{-6}$   & 0.143 \\  
\hline  
\end{tabular}  
\end{adjustbox}
\caption{\textbf{Two-way repeated-measures ANOVA on error rates}, reporting SS: Sum of Squares, DF: Degrees of Freedom, MS: Mean Square; F: F-statistic, uncorrected p-values ($p$-unc), and partial effect sizes ($\eta_p^2$). Input structure shows the largest effect, followed by Model and their interaction; all are statistically significant.}  
\label{tab:anova_number_errors}  
\end{table*}  

%% file: latex/tables/name_anova.tex
\begin{table*}[htbp]  
\centering  
\begin{adjustbox}{max width=\textwidth}
\begin{tabular}{lcccccc}  
\hline  
\textbf{Source} & \textbf{SS} & \textbf{DF} & \textbf{MS} & \textbf{F} & \textbf{$p$-unc} & $\eta_p^2$ \\  
\hline
Input Structure            & 23.15 & 2   & 11.57 & 73.48 & $7.35\times10^{-24}$  & 0.458 \\  
Model                      & 0.78  & 1   & 0.78  & 4.98  & 0.027  & 0.028 \\  
Model $\times$ Input Structure & 0.22  & 2   & 0.11  & 0.71  & 0.493  & 0.008 \\  
\hline  
\end{tabular}  
\end{adjustbox}
\caption{\textbf{Two-way repeated-measures ANOVA on name error rates}, reporting SS: Sum of Squares, DF: Degrees of Freedom, MS: Mean Square; F: F-statistic, uncorrected p-values ($p$-unc) and partial eta-squared ($\eta_p^2$). Input structure shows a large, significant effect; model is small but significant; the interaction is not significant.}  
\label{tab:anova_name_errors}  
\end{table*}  

%% file: latex/tables/word_objective_anova.tex
\begin{table*}[htbp]  
\centering  
\begin{adjustbox}{max width=\textwidth}
\begin{tabular}{lcccccc}  
\hline  
\textbf{Source} & \textbf{SS} & \textbf{DF} & \textbf{MS} & \textbf{F} & \textbf{$p$-unc} & $\eta_p^2$ \\  
\hline  
Input Structure            & 28.35 & 2   & 14.17 & 69.11 & $8.14\times10^{-23}$  & 0.443 \\  
Model                      & 0.37  & 1   & 0.37  & 1.79  & 0.183  & 0.010 \\  
Model $\times$ Input Structure & 0.13  & 2   & 0.06  & 0.31  & 0.737  & 0.004 \\  
\hline  
\end{tabular}  
\end{adjustbox}
\caption{\textbf{Two-way repeated-measures ANOVA on word objective error rates}, reporting SS: Sum of Squares, DF: Degrees of Freedom, MS: Mean Square; F: F-statistic, uncorrected p-values ($p$-unc) and partial eta-squared ($\eta_p^2$). Input structure shows a large, significant effect; model and the interaction are not significant with negligible effect sizes.}
\label{tab:anova_word_objective_errors}  
\end{table*}  

%% file: latex/tables/word_subjective_anova.tex
\begin{table*}[htbp]
\centering
\begin{adjustbox}{max width=\textwidth}
\begin{tabular}{lcccccc}
\hline
\textbf{Source}                  & \textbf{SS} & \textbf{DF} & \textbf{MS}
                                 & \textbf{F}  & \textbf{$p$-unc}
                                 & \(\eta_p^2\) \\
\hline
Input Structure                  & 7.88        & 2           & 3.94
                                 & 44.61       & $2.29\times10^{-16}$
                                 & 0.339        \\
Model                            & 3.91        & 1           & 3.91
                                 & 44.34       & $3.48\times10^{-10}$
                                 & 0.203        \\
Model \(\times\) Input Structure & 0.17        & 2           & 0.09
                                 &  0.99       & $0.375$
                                 & 0.011        \\
\hline
\end{tabular}
\end{adjustbox}
\caption{\textbf{Two-way repeated-measures ANOVA on word-subjective error rates}, reporting SS: Sum of Squares, DF: Degrees of Freedom, MS: Mean Square; F: F-statistic, uncorrected p-values ($p$-unc) and partial eta-squared ($\eta_p^2$). Input structure and model show significant effects with moderate effect sizes; the interaction is not significant with a negligible effect size.}

\label{tab:anova_word_subjective_errors}
\end{table*}

%% file: latex/tables/context_anova.tex
\begin{table*}[htbp]  
\centering  
\begin{adjustbox}{max width=\textwidth}
\begin{tabular}{lcccccc}  
\hline  
\textbf{Source} & \textbf{SS} & \textbf{DF} & \textbf{MS} & \textbf{F} & \textbf{$p$-unc} & $\eta_p^2$ \\  
\hline  

Input Structure            & 8.02  & 2   & 4.01  & 54.20 & $5.03\times10^{-19}$  & 0.384 \\  
Model                      & 0.30  & 1   & 0.30  & 4.01  & 0.047  & 0.023 \\  
Model $\times$ Input Structure & 0.59  & 2   & 0.30  & 4.01  & 0.020  & 0.044 \\  
\hline  
\end{tabular} 
\end{adjustbox}
\caption{\textbf{Two-way repeated-measures ANOVA on context error rates}, reporting SS: Sum of Squares, DF: Degrees of Freedom, MS: Mean Square; F: F-statistic, uncorrected p-values ($p$-unc) and partial eta-squared ($\eta_p^2$). Input structure shows a large, significant effect; model is small but significant; the interaction is also significant with a small effect size.}  
\label{tab:anova_context_errors}  
\end{table*}  

%% file: latex/tables/example_annotations.tex
\begin{figure*}[!htb]
{\small
    \lineacross{}
    \begin{center}
    \textbf{Partial-game summary from Qwen-Unstructured input}
  \end{center}

     \vspace{2mm}
    \markerrname{Jalen Green's} two \markerrwordobj{free throws} brought the Rockets within \markerrnumber{12}-\markerrnumber{8}, and a subsequent 3-pointer by \markerrcontext{Frank Ntilikina} \markerrwordobj{tied the game at} \markerrnumber{14}-\markerrnumber{14}. The quarter ended with Cleveland \markerrwordobj{leading} 32-\markerrnumber{36}, thanks to a \markerrwordobj{flurry of 3-pointers from both sides}, including a buzzer-beater by \markerrcontext{Christian Wood}, who hit a 3-pointer to cap off a strong first quarter for Houston.\\
   \textbf{ List of errors:}
    \vspace{-3mm}
    \begin{itemize}
    \item \markerrname{Jalen Green's}: incorrect name, it should be `Cam Whitmore.'
    \item \markerrwordobj{free throws}: incorrect action (word objective), the player made `two layups.'
    \item \markerrnumber{12}-\markerrnumber{8}: incorrect scores (number), the correct numbers/scores are `25-22.'
    \item \markerrcontext{Frank Ntilikina}: incorrect context, he did not play in the NBA this season (2025); `Fred VanVleet' is the correct player.
    \item \markerrwordobj{tied the game at}: incorrect wording, he `cut the deficit to' with the mentioned 3-pointer. 
    \item \markerrnumber{14}-\markerrnumber{14}: incorrect scores (number), the correct numbers/scores are `27-25.'
    \item \markerrwordobj{leading}: incorrect wording, Cleveland was `trailing.'
    \item \markerrnumber{36}: incorrect score (number), it should be `39.'
    \item \markerrwordobj{flurry of 3-pointers from both sides}: word objective error, there were only a couple of three-pointers by Rockets.
    \item \markerrcontext{Christian Wood}: incorrect context, he did not play for Rockets or NBA in the 2025 season. `Cam Whitmore' is the correct player. 
    \end{itemize}
    \begin{center}
    \textbf{Partial-game summary from Qwen-Row input} 
    \end{center}
    The Lakers' bench \markerrwordsub{provided a spark}, with \markerrname{Max Christie} and Shake Milton combining for a series of clutch plays. A critical turnover by \markerrname{Deandre Ayton} and a subsequent three-pointer from Austin Reaves extended the Lakers' lead to 88-75 as the quarter ended. \\
    
   \textbf{ List of errors:}
    \vspace{-3mm}
    \begin{itemize}
    \item \markerrwordsub{provided a spark}: subjective phrase, they combined for 6 points at the end of Q3. Some may agree scoring over 10 points to be spark.
    \item \markerrname{Max Christie}: he was not a bench player and he did not contribute during the clutch play of that quarter. The right bench player is `Christian Koloko.'
    \item \markerrname{Deandre Ayton}: incorrect name, Turnover was made by `Deni Avdija.'

    \end{itemize}


    \begin{center}
    \textbf{Partial-game summary from Llama-JSON input} 
    \end{center}
    The fourth quarter saw the Trail Blazers make a late push, with Anfernee Simons making a 3-pointer and Scoot Henderson adding a jump shot to cut the deficit to 94-86. However, the Lakers were able to hold them off, with LeBron James making a \markerrwordobj{layup} and Max Christie adding a dunk to give them a 103-97 lead. \\

    \textbf{ List of errors:}
    \vspace{-3mm}
    \begin{itemize}
    \item \markerrwordobj{layup}: word objective error, he made a jump shot.
    \end{itemize}
    \caption{\textbf{More examples of partial game summaries with error annotations} to demonstrate errors faced in different input structures (Row, JSON and Unstructured). Corrections are included here for clarity. \emph{Notation:} superscripts N, U, WO, WS, C, O, and X indicate the error categories: Name, Number, Word-objective, Word-subjective, Context, Other, and Not-checkable; this is used here for readability; see the Annotation Protocol \Cref{subsec: annotation} for definitions.}
    \lineacross{}
    
    \label{fig:example_annotations}
}
\end{figure*}


%% file: latex/tables/prompt.tex
\begin{figure*}[ht]
\centering
\fbox{%
\begin{minipage}{0.93\textwidth}
\textbf{NBA Play-by-Play Game Summary Prompt} \\[6pt]

You are a professional basketball reporter hired by my company to cover NBA games. You are provided with a hierarchical JSON play-by-play dataset, which includes structured team information and a chronological list of plays. Each play entry contains key details such as time remaining, the team involved, play description, event type, player names, and score updates. \\[6pt]

Your task is to analyse the data and identify the four most significant moments in each quarter, key scoring plays, momentum shifts, lead changes, game ties, and clutch performances. Prioritise impactful events such as three-pointers, layups, dunks, free throws, and possessions immediately following turnovers or rebounds. Ignore minor plays that do not significantly influence the game’s flow. \\[6pt]

Write a fluent, engaging, and objective game summary of exactly 450 words, structured around these defining moments. The narrative should clearly convey why these moments were significant, seamlessly integrating them into the game’s overall storyline. It should flow naturally from start to finish, capturing the game's pace and intensity. \\[6pt]

Mention the final score and quarter-ending scores for context, but do not rigidly structure the summary by time. Discuss key players in relation to the game-defining plays, without calculating total points. Maintain a smooth, journalistic writing style. Do not write in bullet points, lists, or unnecessary formatting. The summary must be factually accurate, strictly based on the provided data, and completely free from external assumptions, while maintaining a compelling and professional tone. \\[6pt]

Quarter-End Context: Mention the total score contextually within the narrative as each quarter concludes. DO NOT list the quarter scores separately at the very end of the summary.
\end{minipage}
}
\caption{\textbf{Prompt used for generating NBA game summaries from hierarchical JSON play-by-play data.} The first paragraph specifies the input format, while the remaining instructions are consistent across all runs.}
\label{fig:game_summary_prompt}
\end{figure*}